\pdfoutput=1

\documentclass[11pt]{article}

\usepackage[]{ACL2023}

\usepackage{times}
\usepackage{latexsym}

\usepackage[T1]{fontenc}

\usepackage{booktabs}
\usepackage{tabularx}
\usepackage[textsize=scriptsize]{todonotes}
\usepackage{multirow}
\usepackage{latexsym}
\usepackage{pifont}

\usepackage[a-1b]{pdfx}

\usepackage[inline]{enumitem}

\usepackage{tikz}
\usetikzlibrary{positioning}
\usepackage{url}
\usepackage{amsmath}
\usepackage{mathtools}
\usepackage{amssymb}
\usepackage{amsthm}
\usepackage{tablefootnote}
\usepackage{caption}
\usepackage{subcaption}
\usepackage{bbm}
\usepackage{comment}
\usepackage{xspace}
\usepackage{proof}
\usepackage{stmaryrd}
\usepackage{fdsymbol}
\usepackage{bm}
\usepackage{pifont}
\usepackage{algorithm}
\usepackage{algpseudocode}
\usepackage{color, colortbl}
\usepackage{float}
\usepackage{scrextend}

\usepackage[utf8]{inputenc}

\usepackage{microtype}

\usepackage{inconsolata}

\DeclareMathOperator*{\argmax}{arg\,max}

\title{EEL: Efficiently Encoding Lattices for Reranking}

\author{Prasann Singhal$^\diamondsuit$ \quad\quad Jiacheng Xu$^\spadesuit$ \quad\quad Xi Ye$^\diamondsuit$ \quad\quad Greg Durrett$^\diamondsuit$ \\
\textsuperscript{$\diamondsuit$}The University of Texas at Austin, \textsuperscript{$\spadesuit$}Salesforce AI \\
    \texttt{ \{prasanns, xiye, gdurrett\}@cs.utexas.edu, jiacheng.xu@salesforce.com}}

\newcommand{\STAB}[1]{\begin{tabular}{@{}c@{}}#1\end{tabular}}
\newtheorem{definition}{Definition}
\newcommand\sbullet[1][.5]{\mathbin{\vcenter{\hbox{\scalebox{#1}{$\bullet$}}}}}

\definecolor{LightCyan}{rgb}{0.88,1,1}
\definecolor{White}{rgb}{1,1,1}

\begin{document} 

\maketitle
\begin{abstract}
Standard decoding approaches for conditional text generation tasks typically search for an output hypothesis with high model probability, but this may not yield the best hypothesis according to human judgments of quality. Reranking to optimize for \emph{downstream} metrics can better optimize for quality, but many metrics of interest are computed with pre-trained language models, which are slow to apply to large numbers of hypotheses. We explore an approach for reranking hypotheses by using Transformers to efficiently encode lattices of generated outputs, a method we call EEL. With a single Transformer pass over the entire lattice, we can approximately compute a contextualized representation of each token as if it were only part of a single hypothesis in isolation. We combine this approach with a new class of \emph{token-factored rerankers} (TFRs) that allow for efficient extraction of high reranker-scoring hypotheses from the lattice. Empirically, our approach incurs minimal degradation error compared to the exponentially slower approach of encoding each hypothesis individually. When applying EEL with TFRs across three text generation tasks, our results show both substantial speedup compared to naive reranking and often better performance on downstream metrics than comparable approaches.\footnote{Code available at \url{https://github.com/PrasannS/eel-reranking}.}
\end{abstract}

\section{Introduction}

\begin{figure}[t!]
\centering
\includegraphics[width=0.95\linewidth, trim=10 580 1020 10,clip]{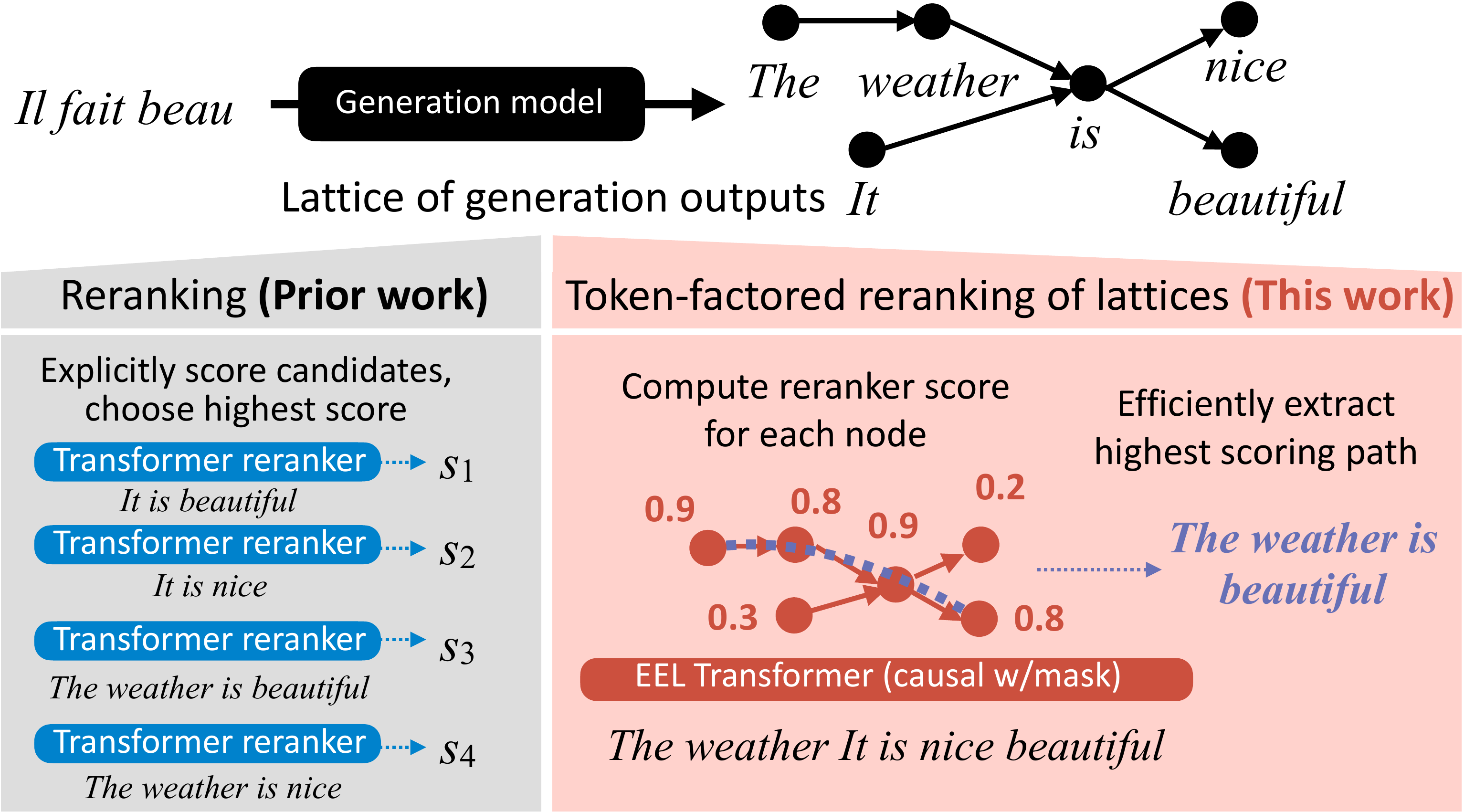}

\caption{Overview of our approach. For a task such as translation, we can generate a lattice of plausible model outputs. Reranking these outputs with Transformers can be slow; we can achieve speedups by efficiently encoding the lattice in a single Transformer pass and using a token-factored reranker to efficiently find the best hypothesis in the lattice.} 
\label{fig:overview}
\end{figure}

Part of the progress in natural language generation over the past few years has been driven by a proliferation of decoding techniques, from beam search to sampling approaches like nucleus sampling \cite{holtzman-nucleus-sampling}, typical decoding \cite{meister-typical-decoding-2022}, and contrastive decoding \cite{li-constrastive-decoding-22}. These techniques, however, only optimize for probabilistic objectives, rather than alignment with human judgments, which is typically better encapsulated by \emph{downstream metrics} \cite{zhang2019bertscore,dhingra-etal-2019-handling,sellam-etal-2020-bleurt,rei-etal-2020-comet} that specifically estimate human preference. Transformer \cite{vaswani2017attention} based \emph{rerankers}, that assign estimated downstream scores to generation candidates, have recently made inroads in translation \cite{lee-etal-2021-discriminative,bhattacharyya-etal-2021-energy,rei-etal-2021-references,freitag-etal-2022-high,fernandes-etal-2022-quality}, open-ended generation \cite{krishna-etal-2022-rankgen}, and summarization \cite{ravaut-etal-2022-summareranker,song-etal-2021-new}. 

However, using rerankers poses several practical challenges. Rerankers work best over a large number of candidates, but generating large sets through beam search is slow. Recent work \cite{xu-etal-2022-massive} has demonstrated the potential to derive and represent large candidate sets in directed acyclic graphs called \emph{lattices}, but the problem remains that naively reranking these large sets is infeasible: scoring each candidate requires one, or even multiple \cite{fernandes-etal-2022-quality} Transformer inference calls. Classic approaches for searching in lattices effectively \cite[inter alia]{koehn-2004-pharaoh,dyer-resnik-2010-context} do not apply to Transformer rerankers, and there is no previously known approach for efficiently extracting good candidates from lattices. This paper proposes an approach to do exactly that, even on lattices encoding thousands of candidates.

We first propose a new class of reranker, the \emph{token-factored reranker} (TFR), that allows efficient inference over a lattice by enforcing a causal mask and decomposing metric scores to the token level, allowing for flexible and efficient scoring while still performing at the same level as standard rerankers. We then show that lattices of generated hypotheses can be efficiently encoded by a Transformer in a single pass by using a custom attention mask and modified position encodings. We call this technique \emph{EEL: Efficient Encoding of Lattices}. EEL enables fast TFR encoding of a large set of generation outputs at once, specifically enabling rapid extraction of the hypothesis with the highest TFR score; see Figure~\ref{fig:overview}.

We evaluate our approach on lattices constructed from beam search as well as lattice decoding. Across three generation tasks (machine translation, summarization, and table-to-text generation), with several downstream metrics, we show that EEL is able to find optimal candidates with respect to the TFR with minimal degradation compared to an exhaustive approach. That is, the highest-scoring candidate from the efficient encoding is nearly as high quality as the highest scoring candidate from naively encoding all candidates independently. Moreover, we show that our approach is efficient and leads to gains on downstream metrics compared to naive reranking approaches. We further propose a method to diversely sample multiple varying hypotheses while reranking with respect to scoring metrics, allowing us to identify different optimal ``modes'' within an input lattice. Our approaches are particularly effective when used with lattice decoding, though we also demonstrate substantial speedups when reranking beam search outputs. 

\textbf{Our Contributions: } (1) We introduce a new class of reranker, the token-factored reranker (TFR), that can support efficient inference over lattices. (2) We propose a method for encoding a lattice with a Transformer (EEL) that enables efficient reranking with TFRs with minimal degradation compared to exhaustive search. (3) We compare beam search, lattice decoding, and multiple reranking strategies across three NLG problems. 

\section{Setting}

Our approach centers on reranking output from conditional text generation with neural models \cite{sutskever2014sequence,bahdanau2014neural}. Such models place distributions over target output $\mathbf{y} = (y_1,\ldots,y_n)$ given input sequence $\mathbf{x}$ via a text generation model $\theta$: $p(\mathbf{y} \mid \mathbf{x}; \theta) = \prod_{k=1}^n p(y_{k} \mid \mathbf{y}_{<k}, \mathbf{x}; \theta)$. In trying to optimize this model probability $p(\mathbf{y} \mid \mathbf{x};\theta)$ by finding the most likely $\mathbf{x}$, decoding algorithms often produce \textbf{candidate sets} $H$ of highly likely outputs under the model for our reranking stage (e.g. for beam search, the candidate set is the N completions for all beams).

Our goal is to rerank candidate sets to optimize for a downstream objective $T(\mathbf{x}, \hat{\mathbf{y}})$. We assume our reranker $S: (\mathbf{x}, \hat{\mathbf{y}}) \rightarrow \mathbb{R}$ scores an (input, hypothesis) pair and returns a real-valued approximation of $T$ using a Transformer model. Reranking can be divided into two steps: (1) the generation of a candidate set $H = \{\hat{\mathbf{y}}^{(1)}, \hat{\mathbf{y}}^{(2)}, \ldots, \hat{\mathbf{y}}^{(M)}\}$, and (2) the extraction of the highest scoring candidate  $\mathbf{y}_{\mathrm{best}} = \argmax_{\mathbf{y} \in H} S(\mathbf{x}, \mathbf{y})$. The end result thus depends on the quality of the candidate set generated by a decoding method as well as how well $S$ approximates $T$. Note, in this paper, we use \emph{reranker} to refer to the model $S$ that assigns scores to hypotheses, which is distinct from the actual reranking procedure itself. 

In our setting, $H$ is represented by a \textit{lattice} which encodes the candidates (further detail in Section~\ref{sec:methodology}). This can either be a packed representation of beam search candidates or can be a more complex lattice generated natively by alternate decoding approaches. Specifically, we use the approach of \citet{xu-etal-2022-massive}, which expands paths with a modified depth first search and merges similar paths in a recombination step to focus search budget on new paths, allowing for generation of larger sets of hypotheses with greater diversity. Notably, these lattices are not tree-structured and contain reentrancies. 

Compared to complete candidate sets of normal text sequences, \emph{lattices can exponentially reduce the number of tokens needed to encode large candidate sets}, enabling strong speedups if leveraged correctly. Thus, step (2) is the focus of this paper: given a scoring model $S$ and a lattice encoding $H$, \textbf{can we encode a lattice and still select the highest scoring candidate encoded in $H$?} Our solution will specialize $S$ to be a \emph{token-factored} reranker, which we define below, and $H$ to be encoded in a lattice; we show that these assumptions hold for strong rerankers that can be applied to real-world generation problems, even when encoding thousands of candidates in as little as a single Transformer pass. We attempt to minimize the error in our selected candidate versus the oracle best candidate $\mathbf{y}_\mathrm{best}$ (defined above), which we refer to as \textit{degradation}.

\section{Reranking Generation Outputs} 

\subsection{Token-factored Rerankers}
\label{sec:tokenfactored}

A key idea in this work is that we can efficiently rerank a lattice encoding of a candidate set given a certain reranker structure. Specifically, if we can decompose the underlying reranking score to be a linear function of tokens in the lattice, we can extract hypotheses efficiently (Section~\ref{sec:reranking_lattices}). We thus propose the \emph{token-factored reranker} (TFR). 


Assume a reranker model $S(\mathbf{x}, \hat{\mathbf{y}})$ that, given a candidate, generates a score evaluating for some downstream objective $T$ (e.g. quality, formality, etc.). Assume moreover that $S$ involves a Transformer $f: (\mathbf{x}, \hat{\mathbf{y}}) \rightarrow \mathbf{h}$ that produces contextualized embeddings $h_i$ for each output token $\hat{y}_i$. These can condition on $\mathbf{x}$, so $f$ can be either an encoder, decoder, or encoder-decoder model. This work primarily examines causally constrained TFRs, defined as follows:

\begin{definition}[token-factored reranker]
\label{def:causal-tfr}
Let $S$ be a \textbf{token-factored reranker} (TFR) if it takes the form $S(\mathbf{x}, \hat{\mathbf{y}}) = \sum_{k=1}^{n} s(f_c(\mathbf{x}, \hat{\mathbf{y}}_{\leq k})_k)$ where $s$ is some linear function and $f_c$ is a \emph{causal} contextualized model that only depends on tokens up to and including $y_k$.
\end{definition}

\noindent
We specialize $f_c$ to be a causal model because we found this important to improve the quality of our approximation. However, theoretically we can also use \textbf{bidirectional token-factored rerankers (bTFRs)} where instead $S(\mathbf{x}, \hat{\mathbf{y}}) = \sum_{k=1}^{n} s(f(\mathbf{x}, \hat{\mathbf{y}})_k)$ for non-causal $f$ (e.g., BERT).

For the aggregation function, we found $s(x) = \frac{x}{n}$, averaging, to work well. 


\paragraph{Generality} TFRs can be trained straightforwardly on any dataset of $(\mathbf{x}, \hat{\mathbf{y}})$ pairs labeled with quality scores. Furthermore, a range of existing architectures fall into or near a TFR-oriented framework. For example, decoding time token-level \emph{model score}, the negative log of the probability of a selected token at a given decoding step, is a TFR. On-the-fly reranking approaches like RankGen \cite{krishna-etal-2022-rankgen} can also factor into a TFR. Furthermore, the sums of two TFRs (or TFRs) will also be usable in our pipeline, allowing us to combine multiple TFRs or use a weighted composition of other token-level scores.

\begin{definition}[ensemble TFR; E-TFR]
\label{def:ensemble-tfr}
Let $S$ be a token-factored reranker. Let $M$ be another TFR where $M(\mathbf{x}, \hat{\mathbf{y}}) = \sum_{k=1}^n \log p(\hat{y}_k \mid \mathbf{x}, \hat{\mathbf{y}}_{<k})$, where $p$ is the probability under the base model. Define the \textbf{ensemble TFR} $S_e(\mathbf{x}, \hat{\mathbf{y}}) = S(\mathbf{x}, \hat{\mathbf{y}}) + \lambda M(\mathbf{x}, \hat{\mathbf{y}})$.
\end{definition}

\noindent
E-TFR ensembles the TFR with model score for better performance in some settings. In this paper, E-TFR specifically refers to this combination ($\lambda=0.75$ in this work), but note that TFRs can be defined or ensembled from any set of models or functions that meet Definition~\ref{def:causal-tfr}.



\subsection{Reranking Lattices}
\label{sec:reranking_lattices}

If we assume that efficient computation of $f_c$ is possible then it becomes easy to optimize $S$ over $H$ when $S$ is a TFR. We can use dynamic programming to extract $\mathbf{y}_{\mathrm{best}}$ from the input lattice.
Algorithm~\ref{alg:bestpath} describes the procedure, which is essentially the Viterbi algorithm with no transition scores. We start with lists \emph{flat} which contains all $v \in V$, sorted topologically, and \emph{ends}, which contains the indices in flat of all nodes in $V$ with no next nodes. We iterate through \emph{flat}, choosing the highest scoring path leading to each node based on whichever previous node has the highest scoring path. Finally, we return the path ending with the highest overall score. In practice it may be necessary to normalize ending scores depending on the function $s$, for example we divide each path score by the length of the path before choosing the best scoring path. 

\begin{algorithm}[t!]
\small
\caption{Extract Best Path in Lattice}\label{alg:bestpath}
\begin{algorithmic}[1]
\Require Topologically sorted list \emph{flat}, list \emph{ends} with all ending nodes $v\in V$, the set of all paths in lattice $P = (v_1, \ldots, v_k)$, $\mathrm{par}(v_i) \subseteq V$ returns set of parents of $v_i$
\Ensure highest scoring path returned
\State $\mathrm{best} \colon V \mapsto (P, \mathbb{R})$
\For{$c \in$ \emph{flat}}
    \State s, $\hat{p}$ $\gets \argmax_{p \in \mathrm{par}(c)}$ {$\mathrm{score}(\mathrm{best}(p))$}
    \State $\mathrm{best}(c)$ $\gets (\hat{p} \cup (c)$, $\mathrm{s}+s(f(c))$) // extend hypothesis
    \State $i \gets i+1$
\EndFor \\
\Return $\argmax_{e \in \emph{ends}} \mathrm{best}(e)$ // return best end state; can extract path via backpointers
\end{algorithmic}
\end{algorithm}

\begin{figure}[ht]
\centering
\includegraphics[width=\linewidth, trim=10 440 2750 10,clip]{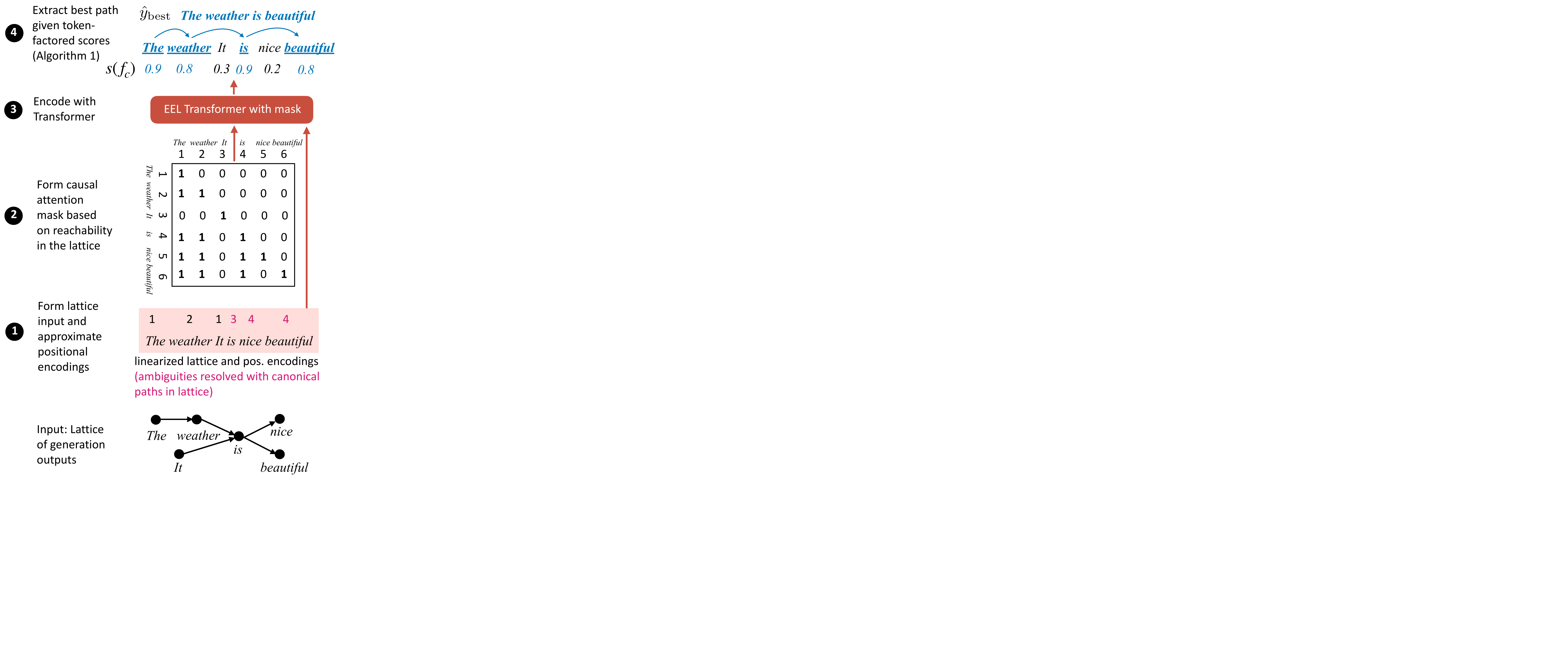}

\caption{Detailed overview of EEL, starting from bottom, using a single context mask and causal reachability.} 
\vspace{-0.16in}
\label{fig:detailed-walkthrough}
\end{figure}

\subsection{Diverse Path Selection}
\label{sec:diverse-path}
In addition to estimating a best path, our approach also supports being able to extract a diverse set of $k$ paths while still optimizing for $S(\hat{\mathbf{y}})$. This can be accomplished by running Algorithm~\ref{alg:bestpath} $k$ times, and at the end of each round, modifying the score of each node $s(y_i) = s(y_i) - w \cdot o(y_i)$, where $w$ is a diversity weight hyper-parameter, and $o(y_i)$ corresponds to the number of times a given token has occurred in previous best path results. 

\label{sec:setting}

\section{Efficient Encoding of Lattices (EEL)}
\label{sec:methodology}

In this section, we discuss the computation of $f_c$: how can we efficiently compute Transformer encodings for a lattice? Across decoding methods, as the total number of lattice nodes is often exponentially fewer than the total number of tokens in $H$, being able to rerank all the candidates by only encoding each lattice token once can lead to a huge reduction of computation. Thus, our goal with EEL is to compute embeddings for all nodes in a set of hypotheses $H$ represented by a directed graph (lattice) $G = (V, E)$ encoding all candidates in $H$. $V$ is the set of all expanded nodes $v_i \in V$, and $E$ the set of directed edges $e_{i, j}$ which represents $v_i \in V$ preceding $v_j \in V$ in some candidate in $H$ (we'll use $v_i$ interchangeably with its associated token $y_i$).\looseness=-1

Figure~\ref{fig:detailed-walkthrough} shows the overall steps of our solution, which we now describe in detail.

\subsection{Constructing Inputs} 

Transformers primarily take three inputs:
\begin{enumerate*}
    \item \textbf{Token Ids}, laid out consecutively in an input canvas;
    \item \textbf{Position Ids}, corresponding to each token's position in text;
    \item \textbf{Attention Mask}, a mask that dictates which tokens can attend to each other.
\end{enumerate*}






To encode a lattice consistently with individual Transformer passes, the tokens each token attends to and the position ids of those tokens should be the same as if just part of a single sequence. As Transformers can only encode one canvas (context window) of tokens at a time, we accordingly need to lay the lattice tokens onto a single token id canvas. For position ids, the default format in most pre-trained Transformers, such as GPT-3, is \textbf{absolute} position ids, where the index of a token $t$ in a sequence of inputs is simply its token id, corresponding directly to its location in the input. 

In order to make lattice position embeddings compatible, we use a \textbf{canonical} position id strategy, where we do a depth-first traversal of all nodes in $G$ with respect to node model probability. Assume $v_i$ is the first node to precede $v_k$ in the traversal and edge $e_{i, k}$ exists; then $pos(v_k) = pos(v_i)+1$ (see Step 1 of Figure~\ref{fig:detailed-walkthrough}). Alternate position id schemes include computing the longest or shortest path from the start \cite{neubig-lattice-attention}, or a random depth-first traversal, though we generally don't note any empirical differences between these approaches. Our method gives slightly better outcomes in certain settings when a lattice doesn't fully fit within a canvas, as it serves as an implicit truncation of low model probability paths.

\subsection{Masking}

To give individual tokens information about their context, Transformer encoders typically allow all input tokens attend to each other. Since our TFR $S$ is trained on normal text data, where a token expects to ``see'' each other position id exactly once, simply passing in linearized lattice inputs leads to large degradation, as tokens that don't share paths can still attend to each other, something models aren't trained to handle. We formulate all masks as $n \times n$ matrices where $n = |V|$ and each index corresponds to a node in $V$ based on position in the canonical canvas. The non-zero values of row $i$ indicate which tokens $y_i$ can attend to. We below walk step-by-step through several mask types to reach a strategy with the lowest degradation (ablations in Table~\ref{tab:method-ablation}): 

\paragraph{Causal Reachability: } We first construct an $n \times n$ adjacency matrix $A$ such $A_{ij} = 1$ if $e_{i,j} \in E$, else 0. We can then obtain a \textbf{causal reachability} mask using the following reachability equation:
\begin{equation}
\label{eq:reachmask}
C = \min(I_{n}+\sum_{i=1}^{l}(A^T)^{i}, \mathbf{1})
\end{equation}
Where $I_n$ is an identity matrix, $l$ is the length of the longest hypothesis in $H$, and the min operation causes all values to be either 0 or 1. Such a mask prevents tokens that aren't in any common hypotheses from attending to each other, but connects a token to all tokens that come before it (we'll call these \emph{contexts}) for all paths in $H$. Note that $A$, and thus $C$ are uni-directional to match the causal formulation of TFRs. 

We can obtain a mask for a bidirectional TFR by replacing $A^T$ in Equation~\ref{eq:reachmask} with $A+A^T$. However, we empirically found that reachability in both directions results in more degradation in the TFR, resulting in our decision to use causal TFRs. Causal TFRs enable lossless encoding for a lattice with no reentrancies. There can be degradation in graphs with reentrancies such as those produced by lattice decoding, due to multiple contexts or mismatched canonical position ids. 

\paragraph{Single Context} To constrain sources of degradation even further, we can limit $A$ to only have a single randomly selected $1$ per row, which would translate to each token only being able to look back at a single path: in other words a \textbf{single context} mask $C^*$. This strategy is equivalent to reranking all hypotheses encoded within a random subgraph $G^* = (V, E^*)$, where $E^* \subseteq E$ such that only one directed edge $e_{i, j}$ exists from any node $v_i$. Thus, remaining degradation is limited to the paths in $G$ not encoded in $G^*$. In Figure~\ref{fig:detailed-walkthrough}, this manifests as \emph{beautiful} not attending to \emph{It}.

\paragraph{Few Mask} We can improve our approximation with higher computational cost using a \textbf{few-mask} variant of EEL. With the same input and position ids, we run the EEL pipeline with ``single context'' with $m$ different starting random adjacency $A$ instances. This allows us to then obtain $m$ different extracted best paths based on initialization, from which we can then choose an overall best scoring path based on the normalized path scores of the $m$ best paths. This allows more potential paths in the lattice to be explored without suffering context-related degradation. Note that while this approach leads to the best performance, single context masking is the best in terms of efficiency. 

\section{Experimental Setup}

\subsection{Settings}

To demonstrate the robustness of our approach, we run experiments on a variety of different base tasks, with lattices generated in different conditions, and with 3 different reranking models that fit our criterion. Our base tasks are as follows:

\paragraph{Machine translation} We generate lattices (using an mBART-large model \cite{mbart-liu-etal}) in 3 machine translation settings from WMT-2019: English to German (\textsc{en-de}), English to Russian (\textsc{En-ru}), and French to English (\textsc{fr-en}) \cite{barrault-etal-2019-findings}.

\paragraph{Table-to-text} We use generated candidate sets from a BART-large model from \citet{wnlg-generation-bart} on examples from the WebNLG 2017 challenge \cite{webnlg2017}.

\paragraph{Document summarization} We also generate a set on document summarization, using a BART-large model \cite{bart-lewis-etal-2020} and XSum data \cite{narayan-etal-2018-dont}.


We experiment with 3 downstream objectives: COMET \cite{rei-etal-2020-comet} quality estimation of machine translations, PARENT score \cite{dhingra-etal-2019-handling} precision for quality estimation of table-to-text generation, and number of unique nouns\footnote{We choose this somewhat synthetic setting to provide a reranking setting for summarization and to show that EEL works on diverse types of downstream metrics, not just continuous scores from 0-1.} using a part-of-speech tagger. We train TFRs for each respectively; see Section~\ref{subsec:models} for model details. 

\paragraph{Implementation Details} We generate sets of 500 lattices each with different generation conditions, across 5 settings (3 MT settings, summarization, table-to-text), parallel for 3 decoding methods: 

\begin{addmargin}[1em]{0em}

$\sbullet[0.75]$ \textbf{lattice decoding} (\textsc{latt}), with a beam width of 4\footnote{\citet{xu-etal-2022-massive} call this an ``equivalent beam size.'' We run with a lower value than our beam search because their method is slower than beam search due to poorer parallelism.} and RCB recombination (based on n-gram matching during decoding) \cite{xu-etal-2022-massive}.

\noindent $\sbullet[0.75]$ beam search, with \textbf{beam-width 12} (\textsc{b-12}), as a low generation cost baseline

\noindent$\sbullet[0.75]$ beam search, with \textbf{beam-width 50} (\textsc{b-50}), which we find to have a comparable wall clock generation time to \textsc{latt}

\end{addmargin}

For beam search, we use the Hugging Face generate() API to generate candidates, and we use the corresponding model probabilities returned alongside generation candidates as model scores.

\subsection{TFR Training}
\label{subsec:models}
We fine-tune three TFR (Section~\ref{sec:tokenfactored}) models.

    \paragraph{MT-TFR} Downstream objective: COMET score, for reference-based machine translation quality estimation. MT-TFR uses an XLM-RoBERTa-Base \cite{xlm-roberta} encoder to encode both source and hypothesis sentences, and is fine-tuned on COMET scores generated on the multi-lingual WMT17-21 direct assessment sets \cite{wmt-2021-machine}. Note that we are estimating a reference-based metric in a reference-free way similar to COMET-QE \cite{rei-etal-2021-references}.
   \paragraph{TAB-TFR} Downstream objective: PARENT precision, for reference-based table-to-text generation quality estimation. We generate 16 candidates using a T5-large generation model \cite{wang-etal-2021-stagewise} for each examples in the WebNLG 2017 training set, and obtain PARENT precision scores for each of these to acquire our training set labels. For the TFR encoder, we use a BART-Base encoder-decoder model, using the encoder to encode the source, and the decoder to encode the candidates. 
    \paragraph{NOUN-TFR} Downstream objective: number of unique nouns. The model is trained using an XLM-RoBERTa-Base \cite{xlm-roberta} encoder on a set of 100,000 English sentences from the news-test-14 dataset. We fine-tune it to predict how many unique tokens with the NN, NNS, or NNP POS tags are passed into the candidate sentence, using an NLTK \cite{nltk} POS tagger as our gold labeler (we normalize this by dividing by 10). Note that, while we believe Noun-TFR correlates with diversity and informativeness, as more nouns may indicate more specific content around entities, we are not setting it up as a gold standard for summarization quality; it's designed more as a testbed for our approach. 

\subsection{TFRs vs Non-TFR Rerankers}

To confirm that token-factored rerankers do not have worse downstream performance, we validate TFR models against COMET-QE \cite{rei-etal-2021-references}, a non-token factored reranking metric. When refactoring COMET-QE to be token-factored, and fine-tuning, we're able to reproduce reranking performance when ensembled with model score (see Section~\ref{sec:down-setup}): 0.699 (French to English), 0.598 (English to German), and 0.650 (English to Russian) respectively (see more in \ref{appendix:tfr-validation}). With a TFR model, we find downstream performance to be 0.698 (French to English), 0.576 (English to German), and 0.614 (English to Russian); on average, only 0.02 COMET score worse. In other words, \textbf{TFRs don't significantly fall behind other Transformer-based reranking models}.

\definecolor{OracleColor}{rgb}{0.88,1,1}
\newcommand{\oc}[0]{\cellcolor{OracleColor}}

\begin{table*}[ht]
    \centering
    \renewcommand{\tabcolsep}{1.2mm} 
    \footnotesize
    \scalebox{0.9}{
    \begin{tabular}{cl|ccc|cc|c|ccc}
            \toprule \multicolumn{2}{l|}{}

                      &  \multicolumn{3}{c|}{MT-TFR} & \multicolumn{2}{c|}{NOUN-TFR} & \multicolumn{1}{c|}{TAB-TFR} & \multicolumn{3}{c}{Efficiency (Fr-En)}\\
        \multicolumn{2}{l|}{} &  \multicolumn{3}{c|}{reranker score} &  \multicolumn{2}{c|}{score}  & {score} &  ratio$\uparrow$  &  \multicolumn{2}{c}{sec $\downarrow$ }  \\
         \multicolumn{2}{l|}{Method} & {\sc fr-en} & {\sc en-de} & {\sc en-ru} & {\sc fr-en} & {\sc xsum} & {\sc webnlg} & {\sc c/n}  & {\sc rrk} & {\sc gen} \\

         \midrule
         \multirow{6}{*}{\STAB{\rotatebox[origin=c]{90}{\sc latt}}} 
         & {\sc rand} & .605 & .582 & .631 & .765 & .831 & .511 & .020 & .051 & \multirow{6}{*}{4.135$\pm$1.595}\\
         &{\sc tfr-8-samp} & .690 & .690 & .792 & .863 & 1.022 & .589 & .020 & .167 \\
         &{\sc tfr-32-samp} & .716 & .719 & .838 & .903 & 1.097 & .627 & .020 & .695 \\
         &{\sc eel 1-mask} & .695 & .700 & .836 & .922 & 1.118 & .653 & \textbf{3.304} & \textbf{.091}\\
         &{\sc eel 8-mask} & \textbf{.720} & \textbf{.731} & \textbf{.862} & \textbf{.934} & \textbf{1.142} & \textbf{.657 }& 0.413 & .252\\
         & \oc {\sc exhaustive} & \oc.743 & \oc.748 & \oc.883 & \oc.945 & \oc1.178  & \oc.692 & \oc.025 & \oc17.950 \\
         \midrule
         \multirow{3}{*}{\STAB{\rotatebox[origin=c]{90}{\sc b-12}}} 
         & {\sc rand} & .629 & .616 & .678 & .751 & .734 & .582 & .024 & 0 & \multirow{3}{*}{1.280$\pm$.260} \\
         & {\sc eel 1-mask} & .687 & .684 & .783 & .812 & .848 & .641 & \textbf{.064} & \textbf{.078} \\
          & \oc{\sc exhaustive} & \oc.687 & \oc.684 & \oc.783 & \oc.812 & \oc.848 & \oc.641 & \oc.024 & \oc.248 \\
         \midrule
          \multirow{3}{*}{\STAB{\rotatebox[origin=c]{90}{\sc b-50}}}
         & {\sc rand} & .618 & .611 & .640 & .752 & .733 & .581 & .025 & 0 &  \multirow{3}{*}{3.670$\pm$.960} \\
         & {\sc eel 1-mask} & .700 & .707 & .805 & .845 & .908& .651 & \textbf{.075} & \textbf{.120} \\
         & \oc {\sc exhaustive} & \oc .706 & \oc.710 &  \oc.810 & \oc.850 & \oc.909 & \oc.653 & \oc.025 & \oc 1.051 \\
         \bottomrule
    \end{tabular}
    }
    \caption{Base task results, grouped by decoding method of input lattices. \textsc{EEL 1-mask} and \textsc{EEL 8-mask} strongly improve candidate/node (\textsc{c/n}) efficiency and demonstrate notable speedups compared to baselines. Across settings \textsc{EEL 8-mask} comes close to matching the much slower \textsc{latt} exhaustive, even outperforming \textsc{b-50} exhaustive, demonstrating that \textsc{EEL} comes with low degradation for high efficiency gain. }
    \label{tab:main-tfr}
    \vspace{-0.15in}
\end{table*}

\subsection{Evaluating Efficiency}


\paragraph{Wall clock time: } The main costs in a reranking system are the \emph{generation time} (\textsc{gen}) of the decoding method, and the \emph{reranking time} (\textsc{rrk}) to process and rerank the input. We report these in the form of wall clock time as a practical heuristic. 

\paragraph{Candidates / Nodes: } To control for differences in architecture, parallelism, and other factors that influence wall-clock time, we also report a measure we call \emph{candidates per nodes} (\textsc{c/n}). Specifically, for a given approach, we measure how many candidates it encodes, and the number of nodes (tokens) the TFR needs to rerank the candidate set. 

\section{Intrinsic Evaluation: Finding Hypotheses with High Reranker Scores}
\label{sec:intrinsic}

We first evaluate how closely the TFR scores of \textsc{eel} selections come to matching the quality of the oracle top-1 hypotheses $\mathbf{y}_{\mathrm{best}}$ with respect to TFR model $S(\mathbf{x}, \hat{\mathbf{y}})$. We compute an upper bound for this by calculating TFR scores on \textbf{exploded} lattice candidate sets: enumerating the exponentially-large set of complete paths and scoring each individually. Our \textsc{exhaustive} numbers are the average top-1 TFR score (not the downstream metric score) across the entire exploded set.

As baselines, we rerank on a randomly sampled sets of 1 (\textsc{rand}), 8 (\textsc{tfr-8-samp}), and 32 (\textsc{tfr-32-samp}) from exploded sets. For \textsc{eel} approaches, we test a single mask (\textsc{eel 1-mask}) and few-mask (\textsc{eel 8-mask}) approach. We report these results across decoding methods in Table~\ref{tab:main-tfr} alongside efficiency measurements on the \textsc{MT-TFR} \textsc{FR-EN} setting, which we noted to be representative of performance across settings.

\paragraph{EEL universally provides strong efficiency boosts:} While the lattice \textsc{exhaustive} reranking always performs the best, it's incredibly slow, taking 17.95s on average to rerank a single candidate set. \textsc{EEL 1-mask}, meanwhile, encodes the same candidate set roughly \textbf{200 times faster}, at only .091s, and even \textsc{EEL 8-mask} only takes .252s. The candidate / node ratio of 3.304 (\textsc{EEL 1-mask}) and .413 \textsc{EEL 1-mask}, compared to baseline \textsc{c/n} efficiencies of ~.025, further demonstrates how much computation \textsc{EEL} saves on large lattices. Even on \textsc{b-50} and \textsc{b-12}, we see strong reranking efficiency improvements, with 3x and 2.67x better \textsc{c/n} efficiency, and notable rerank time (\textsc{rrk}) boosts. 

\paragraph{EEL selections nearly match \textsc{oracle}: } While \textsc{eel}, especially with large input lattices, can improve encoding efficiency by orders of magnitude, it does so while on average still coming very close to matching oracle top-1 candidates (see Appendix~\ref{appendix:degradation} for degradation analysis). We find \textsc{eel-8-mask} on \textsc{latt} to be the most effective overall for efficiently approximating lattice decoding lattice TFR score, as outside of the unfeasible \textsc{latt} \textsc{exhaustive}, \textbf{\textsc{eel 8-mask} obtains the highest TFR scores in every setting}, outperforming baselines and even \textsc{b-50} \textsc{exhaustive}. Furthermore, \textsc{eel}, applied on \textsc{b-12}, and \textsc{b-50} comes with zero and near-zero degradation respectively. 

\begin{table}[t]
    \centering
    \renewcommand{\tabcolsep}{0.9mm}
    \footnotesize
    \begin{tabular}{cl|ccc|c|c}
            \toprule \multicolumn{2}{l|}{}
           & \multicolumn{3}{c|}{COMET} & \multicolumn{1}{c|}{NOUN} & \multicolumn{1}{c}{PRT-P} \\
         \multicolumn{2}{l|}{Method} & {\sc fr-en} & {\sc en-de} & {\sc en-ru} & {\sc xsum} & {\sc wnlg} \\
         \midrule
         & {\sc rand-b50}  & .654 & .541  & .482 &  7.01 & .573 \\
         & {\sc rand-latt}  & .598 & .419  & .445 & 8.07 & .452\\
         & {\sc b50-prob} & .680 & .564  & .518 & $-$ & .623 \\
         &{\sc latt-prob}  & .660 & .493 & .545 & $-$ & .654 \\
         \midrule
         
         &{\sc eel-w 1-mask} & .673 & .541  & .592* &  10.80* & .667\\
         &{\sc eel-w 8-mask}  & .674 & .545  & \textbf{.593*} & \textbf{11.05*} & \textbf{.670} \\
         & {\sc b50-e-tfr}  & \textbf{.689} & \textbf{.576} & .562 &  8.63 & .664\\
         \midrule
         \multicolumn{7}{c}{Oracle values}\\ \midrule
         &{\sc latt-e-tfr}  & .698 & .574  & .614 & 11.24 & .691 \\
         &{\sc b50-oracle}  & .761 & .664 & .702  & 8.74 & .778 \\
         &{\sc latt-oracle}  & .789 & .677 & .775 &  11.40 & .825\\
         \bottomrule
    \end{tabular}
    \caption{Downstream score results, grouped by systems which have comparable computational requirements. We find that our best method (\textsc{eel-w 8-mask}) achieves strong performance. The bottom rows report oracle reranking approaches, showing that lattices have much higher oracle values and that our rerankers can sometimes come close (especially on NOUN-XSum). * indicates statistically significant improvement over \textsc{b50-e-tfr} using a paired bootstrap test, $p < 0.05$.}
    \vspace{-0.20in}
    \label{tab:downstream}
\end{table}

\section{Downstream Evaluation}
\label{sec:down-setup}

While \textsc{EEL}'s algorithmic objective is to find the best TFR scoring candidate in a lattice, the high-level goal, assuming a good TFR reranker, is to \emph{find a high downstream scoring candidate}. To measure downstream success, we compute results with respect to the original metrics that we distill our TFR models on (COMET, NLTK POS-Tagger, PARENT Precision).

Specifically, we assess whether \textsc{EEL} can enable lattice decoding lattices to efficiently outperform comparable beam search (\textsc{b50}) reranking approaches. For the \textsc{COMET} and \textsc{PRT-P} settings, we use E-TFR, a weighted sum of TFR and token-level model score (\textsc{prob}) (see Section~\ref{sec:tokenfactored}).  

Table~\ref{tab:downstream} reports our results.
In addition to E-TFR \textsc{eel} on lattice decoding (\textsc{eel-w 1-mask}, \textsc{eel-w 8-mask}), we report model score reranking baselines (\textsc{b50-prob}, \textsc{latt-prob}), 
a weighted naive TFR (\textsc{b50-e-tfr, latt-e-tfr}) upper bound, and downstream \textsc{oracle} results. Note, \textsc{oracle} results that need references (\textsc{comet} and \textsc{prt-p}) assume access to human annotation and are unrealistic upper bounds.

\paragraph{\textsc{eel} and TFRs can enable \textsc{latt} to efficiently outperform \textsc{b-50}:} Across settings, the gap between downstream \textsc{e-tfr} performance and \textsc{eel} is minimal: for \textsc{eel 8-mask}, only .024 (\textsc{fr-en}), .029 (\textsc{fr-en}), .019 (\textsc{en-ru}), .019 (\textsc{xsum}), and .021 (\textsc{wnlg}), compared to much larger gaps from random baselines. In other words, \emph{\textsc{EEL} is always quite close to gold TFR reranking selections even on downstream metrics}. In settings where the reranking capabilities of lattice decoding (\textsc{latt-oracle}) strongly outperforms \textsc{oracle} capabilities of \textsc{b50-oracle}, such as \textsc{en-ru} (.073), \textsc{xsum} (2.66), and \textsc{wnlg}(.047), where \textsc{latt-e-tfr} strongly outperforms \textsc{b50-e-tfr}, \textbf{\textsc{eel} on \textsc{latt} likewise outperforms the reranking capabilities of beam search}. Note that, on settings such as \textsc{fr-en} where the oracle gap is a more narrow .028, \textsc{eel} still outperforms \textsc{latt-prob}, but not \textsc{b50}. Recall from Table~\ref{tab:main-tfr} that regardless of decoding algorithm, \textsc{eel} approaches enable us to get the best of reranking with substantially lower computation and time.

\section{Analysis}


\begin{figure}[t!]
\centering
\includegraphics[width=0.85\linewidth]{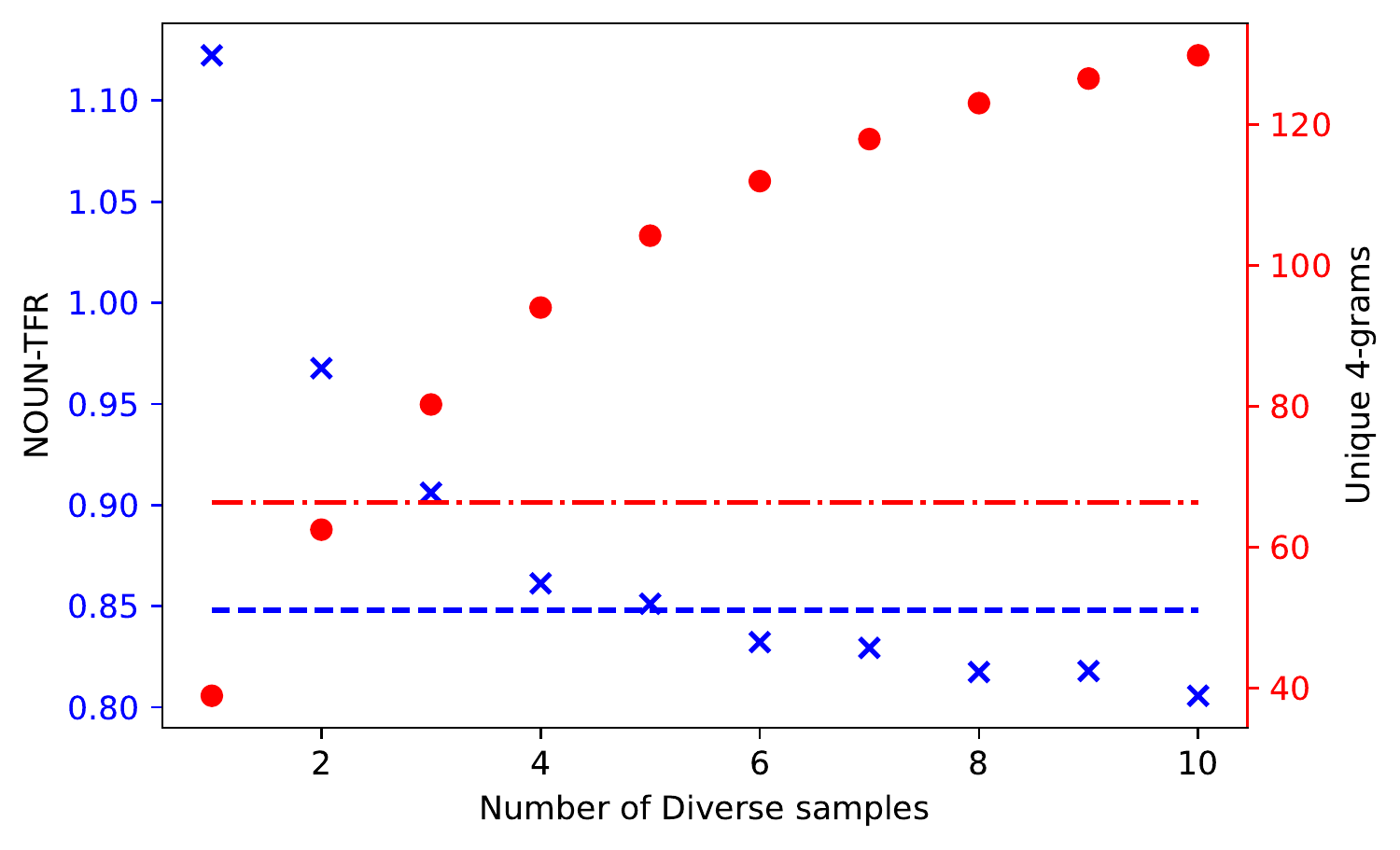}

\caption{Scatter plot, showing, for \textsc{xsum} \textsc{noun-tfr} (blue) and Unique 4-grams (red) for successive diverse samples. Red (Unique 4-grams) and Blue \textsc{oracle noun-tfr} are \textsc{b-12} baselines. Within 3 diverse samples, EEL outperforms the baselines in both NOUN-TFR and Unique 4-grams. } 

\vspace{-0.15in}
\label{fig:diverse-plot}
\end{figure}

\paragraph{Diversity} We further explore whether \textsc{eel} can sample a diverse set of candidates (see Section~\ref{sec:diverse-path}) that still optimize for \textsc{tfr} metrics. We examine this trade-off in Figure~\ref{fig:diverse-plot}, which shows diversity and \textsc{noun-tfr} of diverse samples taken at $n$ steps (x-axis) aggregated across several examples on \textsc{xsum}. While overall diversity increases rapidly, this comes with a trade-off of lower \textsc{TFR} scores, eventually selecting candidates with roughly average \textsc{noun-tfr} score (the random baseline is .831).

That said, we find \textbf{diverse sampling works to produce several diverse samples}. Our approach is able to rapidly obtain a highly diverse set of candidates before the \textsc{noun-tfr} score of samples eventually converges with the random baseline. \textsc{beam-12}, across all 12 candidates, has an average diversity of 66.35 4-grams, and .848 \textsc{oracle} \textsc{noun-tfr}. at 3 samples, we have good diversity and good quality, while the first 5 diverse samples all outperform the \textsc{b-12} \textsc{noun-tfr} and diversity baselines. It's worth noting that the diversity weight is a hyperparameter, and can be adjusted to allow for a slower trade-off. 


\begin{table}[t]
    \centering
    \footnotesize
    \begin{tabular}{cl|c|c}
            \toprule
        \multicolumn{2}{l|}{}
           & \multicolumn{1}{c|}{MT-TFR} & \multicolumn{1}{c}{TAB-TFR}  \\
         \multicolumn{2}{l|}{Method}  &  {\sc Fr-En} &  {\sc Xsum}   \\
         \midrule
         & {\sc Random} & .605 & .831 \\
         & {\sc Oracle} & .743 & 1.178 \\
         \midrule
         & {\sc EEL Full context} & .695 & 1.120 \\
         & {\sc EEL Default pos} & .655 & .989 \\
        \midrule
         \multirow{3}{*}{\STAB{\rotatebox[origin=c]{90}{\sc Multi}}}
         & {\sc eel 1-mask} & .695 & 1.118 \\
         & {\sc eel 8-mask} & .720 & 1.142 \\
         & {\sc eel 16-mask} & .724 & 1.148 \\
         \bottomrule
    \end{tabular}
    \caption{Ablation that validates several choices in \textsc{EEL} pipeline, including posids, contexts, and more masks.}
    \vspace{-0.16in}
    \label{tab:method-ablation}
\end{table}

\paragraph{Ablations} Table~\ref{tab:method-ablation} shows ablations of different parts of our method. We validate that canonical position ids, and single-context masks are necessary for the success of our approach. We further note that while allowing a token to see all valid contexts has similar performance to \textsc{EEL 1-mask}, it doesn't scale to multi-mask versions, and comes with correctness downsides. We also find there to be diminishing returns past 8 mask runs.

\section{Related Work}

\paragraph{Reranking and efficiency} There is substantial past work on reranking, including significant recent work using Transformers for it \cite{rei-etal-2021-references}. Some models can be expressed as TFRs \cite{krishna-etal-2022-rankgen}; however, others like \citet{rei-etal-2021-references} use approaches like pooling of representations of the hypothesis before additional preprocessing. We do not have an approximation for such techniques; we show that TFRs can be successfully trained in this setting, but other approximations beyond TFRs would be a useful contribution of future research. \looseness=-1

Older machine translation work takes advantage of the structure of $n$-gram language models to do ``reranking'' on the fly, either in hierarchical models \cite{forest-reranking} or phrase-based decoding \cite{koehn-2004-pharaoh}. However, recent work on both evaluation \cite{sellam-etal-2020-bleurt, rei-etal-2020-comet} and reranking \cite{rei-etal-2021-references} show how effective Transformers are, suggesting that approximating these is more promising than devising rerankers with inherently more restrictive factorizations.

\paragraph{Background: Lattice Encoding} Work in the speech recognition community \cite{lattention-asr-rescoring, li-parallel-lattice-rescoring} encodes lattices of speech recognition model decoding outputs with LSTMs to decode better candidates in second pass. Other work examines the ability to encode lattices with self-attentional models \cite{neubig-lattice-attention}, for the purpose of augmented text generation \cite{lattice-translation-zhang, lai-etal-2021-lattice}, albeit using models trained on datasets of lattices. As these approaches often require lattice-specific training, and often don't account for degradation, due to alternate, coarser task formulations, they are much more constrained, and not suitable for our setting.

\paragraph{Controlled generation}

While we focus on re-ranking for better generation ``quality'', our work relates to tasks where one may want to re-rank with respect to some control attribute. Prior work examines the ability to adjust output logits specific to controls \cite{yang-klein-2021-fudge}, MCTS \cite{leblond-etal-2021-machine} as well as control classifier guided decoding \cite{dathathri-pplm-2020}.  


\section{Conclusion}

Across a variety of settings, downstream objectives, and decoding methods, we consistently find EEL to provide high quality selections with low degradation and substantial speedup compared to exhaustive top-1 reranking output. We further demonstrate the capability of our approach to select multiple diverse outputs while still optimizing for TFR scores. By proposing a method that can efficiently encode lattices of large candidate sets, through the combination of \textsc{eel} and TFR models, we thus demonstrate the capability of reranking to be more effective and efficient than what was previously possible with naïve approaches.

\section*{Limitations}
 
While our approach is designed to be as broadly applicable as possible, an inherent limitation of our work is that it depends on the usage of causal \textsc{TFR}-style models, which, though architecturally similar to many existing pre-trained models, require hyperparameter search and fine-tuning to replace non-TFRs on downstream tasks. While we find evidence that such models are as capable as other rerankers, and we believe TFRs can be a default design choice going forward with little loss, this extra requirement may still be a barrier for the broad adoption of our approach. 

More broadly, our approach is designed for settings where \emph{some} reranker is available. If it is not possible to train a robust reranker, for example in a low data setting or a setting where evaluation relies entirely on human judgments that cannot be reproduced by models, our approach cannot be applied. However, we believe that the growth in learned functions of human judgments as part of RL from human feedback loops provides a promising avenue to roll out our approach to new settings. 

Our experiments were carefully chosen to represent the capabilities of our models with several base tasks and several reranking objectives. We didn't, however, explore certain domains involving attribute control such as formality or simplicity, choosing instead to explore more quality-based downstream exploration. We showed the applicability of our approach when reranking outputs on languages other than English, but further results on languages with different typological characteristics may show different trends. 

While our work already provides strong speedups both with candidate sets from lattice and beam search decoding, these speedups become even more valuable for approaches that combine multiple rerankers, which have been shown to potentially lead to further improvements in reranking \cite{fernandes-etal-2022-quality}. While we explore this partially in the form of ensembled EEL with model probabilities, more exploration on \textsc{EEL} for multiple rerankers may be valuable.

\section*{Acknowledgments}

This work was supported by NSF CAREER Award IIS-2145280, a grant from Open Philanthropy, a gift from Salesforce, Inc., a gift from Amazon, and a gift from Adobe. Thanks to Darcey Riley, Andr\'{e} Martins, Ben Peters, Ricardo Rei, Ant\'{o}nio Farinhas, Perez Ogayo, and Jos\'{e} Souza for discussion and feedback during the completion of this paper. Thanks as well to the anonymous reviewers for their helpful feedback. 

\bibliography{anthology,custom}
\bibliographystyle{acl_natbib}

\appendix

\section{TFR Model Details}
\label{sec:tfrdetails}
\subsection{Model Architecture}

\label{appendix:tfrarch}
While the over-arching idea of TFRs stays the same, our TFR models all differ slightly in the setup of the final hidden layers. For the \textsc{noun-tfr} model, we actually don't encode the input, as it isn't necessary for the task, and thus, for a given candidate, we only run the feedforward network at the end on individual token hidden states, as opposed to the concatenated vector with the product and difference between it and the pooled input state. 

For the \textsc{mt-tfr} and the \textsc{tab-tfr} model, we follow the output format described in Section~\ref{subsec:models}. The only difference is that for \textsc{MT-TFR}, we use the same encoder model for both the input and the output, and for \textsc{tab-tfr} we follow an encoder-decoder architecture, where the input is acquired by the encoder, and the output by the decoder, both which are fine-tuned separately during training. We do this as the structure of data is more divergent in table-to-text, and thus reasoned that separate encodings of input and output would lead to greater success. 

\subsection{Training}

\label{appendix:tfrtrain}
For training, as a rough heuristic of how well a model would perform as a re-ranker, we optimized for correlation metrics between TFR model scores and downstream metric scores (Pearson, Spearman, and Kendall's Tau correlations). For our MT-TFR validation set, our model reached .879 Pearson, .854 Spearman, and .690 Kendall correlation with COMET score. For our NOUN-TFR model, our model reached .971 Perason, .940 Spearman, and .800 Kendall correlations with gold NLTK noun count. Lastly, for our TAB-TFR model, our model reached .646 Pearson, .632 Spearman, and .470 Kendall correlation. Interestingly, though the correlation metrics weren't good for the TAB-TFR model, the downstream re-ranking still worked well, outperforming model score with similar margins to MT-TFR. We trained each model for an average of roughly 12 hours of training for the TAB-TFR and MT-TFR models, and roughly 3 hours of training for the NOUN-TFR model. 

Note the size of the train / validation set for MT-TFR were 958,122/74,522, for TAB-TFR were 260,668/28,964, and for NOUN-TFR were 90,000/10,001. 

\begin{table}[t]
    \centering
    \small
    \begin{tabular}{l|c|c}
            \toprule 
         \multicolumn{1}{l|}{Model} & Raw Val Size & Val Size \\
         \midrule
         MT EN-DE & 294,497 & 500 \\
         MT FR-EN & 156,121 & 500 \\
         MT EN-RU & 265,807 & 500 \\
         \midrule
         XSum & 11,334 & 500 \\
         \midrule
         WebNLG & 1863 & 500 \\
         \bottomrule
    \end{tabular}
    \caption{Information on evaluation sets used for our results sections. We randomly sample 500 examples to use for our evaluation from different base task datasets to use for our exploration. MT data comes from the WMT-2019 NewsTest Corpus, XSum comes from the XSum test set, and WebNLG from the WebNLG Test Set}
\end{table}

\subsection{Format Validation}
\label{appendix:tfr-validation}
In order to validate that token-factoring doesn't inherently lead to degradations of a model, we test a token-factored version of COMET-QE (2020), a referenceless re-ranking metric. On a validation set, we measure its correlations on randomly sampled WMT Direct Assessments scores to be .479 Spearman, .431 Pearson, and .333 Kendall's Tau correlations. We then modified the model to be token-factored (re-ordered pooling and feed-forward pass), and fine-tuned the model on Direct Assessments data. We found that we were able to reproduce these correlations with .477 Spearman, .431 Pearson, and .333 Kendall's Tau correlations. More-over we measured similar re-ranking performance on a set of beam search 50 candidate sets, thus validating that token-factoring re-rankers doesn't inherently lead to better or worse re-ranking.

\section{Degradation Analysis}

\label{appendix:degradation}

By design, on any token-level task, or task that can be decomposed into a token-level task (we examine reranking), \textsc{EEL} can perform perfectly on lattices without recombination, while offering greater efficiency. This is because the Transformer lattice encoding scheme with causal models in these circumstances is mathematically equivalent to evaluating each path separately: the use of masking and modified positional encodings enables the computation to exactly replicate the standard setting. While we use beam search lattices to demonstrate this, it can apply to a variety of sequence-modeling tasks with Transformers. 

Note the graphs that we look at have 163 (\textsc{b-12}), 551 (\textsc{b-50}), and 431 (\textsc{latt}) nodes on average, with lattice decoding lattices generally having 15+ recombination points, for an average sequence length of around 40 tokens. The lattice-decoding lattices we examine encode between 1000-4000 candidates per lattice .

\paragraph{Sources of remaining degradation} On lattices with recombination, we still experience some degradation. We can pinpoint remaining degradation as observed in results to 2 sources. Firstly, as masks are randomly generated, there's a chance (increasingly small with each new mask) chance that the true top-1 candidate isn't covered. By the nature of our single-context masks, and how they cover every node in a lattice, its often the case that even if the true top-1 path isn't traced by a mask connection, our approach is likely to extract something similar. Additionally, due to recombination, the absolute position ids we encode tokens with will occasionally mismatch at certain points, leading to slightly different scores than if EEL were not used.

\section{Robustness}

\label{appendix:robustness}
\paragraph{TFR-n-samp / EEL} We control for randomness in several of our experiments as follows. For the \textsc{TFR-N-Samp} rows, we sample 10,000 times from each input lattice, averaging those values to get our final results; we note that across robust runs the numbers don't change up to the 3 decimal precision that we report. Likewise, for the \textsc{EEL} numbers, we randomly generate 32 masks, and then sample results 1000 times respectively from those masks. It's further worth noting that past 16 random masks, the 1-best result often repeats from earlier versions (hence the diminishing returns in our ablation study), so we reason this to be a reasonable cut-off. We follow a similar procedure for the downstream table. 

\paragraph{Timing} While the absolute magnitudes of our timing results are subject to variance as a result of varying compute setting and load, we run our timing results several times at different points in time, and note that the relative ranking and overall patterns remain similar across runs. We run these experiments as 10 runs over the entire sets of 500 which we report results on. We apply batching evenly across our approaches and baselines, and we believe that the comparisons are representative of actual performance deltas, as further validated by our \textsc{c/n} results.

\section{Responsible NLP Checklist}

\subsection{Artifacts}

We use several scientific artifacts in this work. 

\paragraph{Data} We use the WMT datasets (2017-2021, licensed under CC-BY-SA-4.0), the WebNLG Dataset (2017, licensed under CC BY-NC-SA 4.0), and the XSUM dataset (MIT License). 

\paragraph{Code} We use open-source code from the COMET repository (APACHE License 2.0), PARENT score \cite{dhingra-etal-2019-handling}, and lattice generation code from the original lattice decoding paper \cite{xu-etal-2022-massive}. 

\paragraph{Generation Models} For generations models we use bart-large-xsum, facebook/mbart-large-50-many-to-one-mmt, facebook/mbart-large-50-one-to-many-mmt, and bart-large fine-tuned on web-nlg data from \cite{wnlg-generation-bart}. 

\subsection{Model Hyperparameters / Infrastructure}

We run all of our experiments on a single PNY NVIDIA Quadro RTX 8000 GPU. 

As we found it to work well across models, for training our TFR models, we use the following common hyperparameters: \textbf{encoder learning rate}: 1e-5, \textbf{learning rate}: 3.1e-5, \textbf{layerwise decay}: 0.9, \textbf{dropout:} 0.1, \textbf{number of encoder frozen epochs}: 0.3, and a \textbf{batch size} of 8. 

We run our NOUN-TFR model for 40000 train steps, our MT-TFR model for 140000 train steps, and our TAB-TFR model for 190000 train steps. This came out to roughly 28 GPU hours total to train the models that we used. Note that beyond manual adjustments, we don't use any sort of hyperparameter grid search. 

We further report another approximately 8 GPU hours for generating and reranking the lattices we used for our evaluation. 

\subsection{Libraries Used}

We use version 3.7 of the NLTK python library \cite{nltk} to extract part of speech tags to use for our NOUN-TFR setting.

\section{Generation Output}

\begin{table*}[t]
    \centering
    \small
    \begin{tabular}{l|c|p{8.5cm}}
            \toprule 
         \multicolumn{1}{l|}{Label} & {\sc comet} & {Text} \\
         \midrule
         Source & -  & Depuis longtemps, plusieurs éléments mettent en péril l'économie américaine :  \\
         Reference & -  & A number of worrying factors about the US economy have been around for a long time:   \\
         \midrule
         E-TFR \#1 & .683 &  For a long time, several factors have threatened the US economy: \\
         E-TFR \#2 & .683 & For a long time, several factors have threatened the US economy: \\
         E-TFR \#3 & .669 & For a long time, several factors have threatened the American economy: \\
         \midrule
         model score rerank \#1 & .420 &  The U.S. economy has long been threatened by several factors: \\
         model score rerank \#2 & .420 & The U.S. economy has long been threatened by several factors: \\
         model score rerank \#3 & .683 & For a long time, several factors have threatened the US economy: \\
         \midrule
         oracle over lattice & .749 & For a long time, a number of factors have been threatening the US economy: \\
         \bottomrule
    \end{tabular}
    \caption{Example 1, French to English, Reranked on \textsc{latt}}
\end{table*}

\begin{table*}[t]
    \centering
    \small
    \begin{tabular}{l|c|p{8.5cm}}
            \toprule 
         \multicolumn{1}{l|}{Label} & {\sc comet} & {Text} \\
         \midrule
         Source & -  & Une enquête d'opinion menée de longue date à travers l'Europe permet de relier les deux.   \\
         Reference & -  & A pan-European opinion survey, which has been carried out for many years, allows us to relate the two.    \\
         \midrule
         E-TFR \#1 & .533 &  A long-standing opinion poll across Europe makes it possible to link the two.\\
         E-TFR \#2 & .549 & A long-term opinion poll across Europe makes it possible to link the two. \\
         E-TFR \#3 & .374 & A long-standing public opinion survey across Europe links the two. \\
         \midrule
         model score rerank \#1 & .207 &   A long-standing opinion poll across Europe links the two. \\
         model score rerank \#2 & .533 & A long-standing opinion poll across Europe makes it possible to link the two. \\
         model score rerank \#3 & .549 & A long-term opinion poll across Europe makes it possible to link the two. \\
         \midrule
         oracle over lattice & .733 & A long-term opinion poll conducted across Europe makes it possible to link these two. \\
         \bottomrule
    \end{tabular}
    \caption{Example 2, French to English, Reranked on \textsc{latt}}
\end{table*}

\begin{table*}[t]
    \centering
    \small
    \begin{tabular}{l|c|p{8.5cm}}
            \toprule 
         \multicolumn{1}{l|}{Label} & {PARENT-P} & {Text} \\
         \midrule
         Source & -  &<H> 250 Delaware Avenue <R> architectural Style <T> Postmodern architecture
 \\
         Reference & -  &  250 Delaware Avenue has the Postmodern style of architecture.  \\
         \midrule
         E-TFR \#1 & .883 &  The architecture style of 250 Delaware Avenue is Postmodern. \\
         E-TFR \#2 & .519 & 250 Delaware Avenue is in the Postmodern architectural style. \\
         E-TFR \#3 & .589 &  250 Delaware Avenue is located in Postmodern architecture style. \\
         \midrule
         model score rerank \#1 & .519 &   250 Delaware Avenue is in the Postmodern architectural style. \\
         model score rerank \#2 & .883 & The architecture style of 250 Delaware Avenue is Postmodern. \\
         model score rerank \#3 & .389 & 250 Delaware Avenue is in the postmodern architectural style. \\
         \midrule
         oracle over lattice & 1.0 & The architectural style of 250 Delaware Avenue is Postmodern. \\
         \bottomrule
    \end{tabular}
    \caption{Example 3, Table to Text, Reranked on \textsc{latt}}
\end{table*}

\begin{table*}[t]
    \centering
    \small
    \begin{tabular}{l|c|p{8.5cm}}
            \toprule 
         \multicolumn{1}{l|}{Label} & {PARENT-P} & {Text} \\
         \midrule
         Source & -  & <H> Bakso <R> ingredient <T> Celery <H> Celery <R> family <T> Apiaceae \\
         Reference & -  &   Celery is a member of the Apiaceae family and is an ingredient of Bakso.\\
         \midrule
         E-TFR \#1 & .857 &  Celery is a member of the Apiaceae family and is an ingredient in Bakso.\\
         E-TFR \#2 & .668 & Celery is part of the Apiaceae family and is an ingredient in Bakso. \\
         E-TFR \#3 & .828 &  Celery is part of the Apiaceae family and is an ingredient of Bakso. \\
         \midrule
         model score rerank \#1 & .447 &    Celery is part of the Apiaceae family and is one of the ingredients in Bakso. \\
         model score rerank \#2 & .857 & Celery is a member of the Apiaceae family and is an ingredient in Bakso. \\
         model score rerank \#3 & .468 & Celery is part of the Apiaceae family and is one of the ingredients of Bakso. \\
         \midrule
         oracle over lattice & 1.0 & Celery is a member of the Apiaceae family and is an ingredient of Bakso. \\
         \bottomrule
    \end{tabular}
    \caption{Example 4, Table to Text, Reranked on \textsc{latt}}
\end{table*}

\begin{table*}[t]
    \centering
    \small
    \begin{tabular}{l|c|p{8.5cm}}
            \toprule 
         \multicolumn{1}{l|}{Label} & {Unique Nouns} & {Text} \\
         \midrule
         Source & -  & The Death of Poor Joe, which dates back to March 1901, was discovered by British Film Institute...\\
         Reference & -  &   The oldest surviving film featuring a Charles Dickens character has been discovered, in the year of the 200th anniversary of the author's birth.\\
         \midrule
         TFR \#1 & 11 &  The earliest known film of Charles Dickens' A Christmas Carol is to be shown in the UK as part of a celebration of the author's bicentenary next year.\\
         TFR \#2 & 11 & The earliest known film of Charles Dickens' A Christmas Carol is to be screened in March as part of a celebration of the author's bicentenary next year.\\
         TFR \#3 & 11 &  The earliest known film of Charles Dickens' A Christmas Carol is to be screened in London as part of a celebration of the author's bicentenary next year. \\
         \midrule
         oracle over lattice & 11 & The earliest known film of Charles Dickens' A Christmas Carol is to be screened in March as part of a bicentenary celebration of the author's work. \\
         \bottomrule
    \end{tabular}
    \caption{Example 5, Summarization, Unique Nouns, Reranked on \textsc{latt}}
\end{table*}

\begin{table*}[t]
    \centering
    \small
    \begin{tabular}{l|c|p{8.5cm}}
            \toprule 
         \multicolumn{1}{l|}{Label} & {Unique Nouns} & {Text} \\
         \midrule
         Source & -  & Regulator Ofcom ruled the performance, by Alexandr Magala of Moldova, was "in line with audience expectations"...\\
         Reference & -  &   ITV show Britain's Got Talent will not be investigated by the broadcasting watchdog over a sword-swallowing act that drew 33 complaints.\\
         \midrule
         TFR \#1 & 13 &  A daredevil sword act on Britain's Got Talent drew no complaints, despite the stunt leaving one contestant fearing for his life, will not be investigated, TV watchdog Ofcom has said \\
         TFR \#2 & 13 & A daredevil sword act on Britain's Got Talent drew no complaints, despite the stunt leaving one contestant fearing for his life, will not be investigated, TV watchdog Ofcom has said.\\
         TFR \#3 & 11 &  A stunt in which a man slid down a pole with a sword lodged in his mouth on Britain's Got Talent will not be investigated, TV watchdog Ofcom has said \\
         \midrule
         oracle over lattice & 13 & A daredevil sword act on Britain's Got Talent will not be investigated over a stunt in which a man fell down a pole with a sword stuck in his mouth, the media watchdog has said. \\
         \bottomrule
    \end{tabular}
    \caption{Example 6, Summarization, Unique Nouns, Reranked on \textsc{latt}}
\end{table*}

\end{document}